\documentclass[10pt,twocolumn,letterpaper]{article}

\usepackage{iccv}
\usepackage{times}

\usepackage[american]{babel}

\usepackage{amsmath, amssymb}
\usepackage{algorithm, algorithmic}
\usepackage[bold]{hhtensor}
\usepackage{multirow}
\usepackage{graphicx}
\usepackage[caption=false,font=footnotesize]{subfig}
\usepackage{url}

\newcommand{\squishlist}{
 \begin{list}{$\bullet$}
  { \setlength{\itemsep}{0pt}
     \setlength{\parsep}{1pt}
     \setlength{\topsep}{1pt}
     \setlength{\partopsep}{0pt}
     \setlength{\leftmargin}{1.5em}
     \setlength{\labelwidth}{1em}
     \setlength{\labelsep}{0.5em} } }

\newcommand{\squishend}{
  \end{list}  }

\graphicspath{{figure/}}
\usepackage[pagebackref=true,breaklinks=true,colorlinks,bookmarks=false]{hyperref}

\iccvfinalcopy 

\def\iccvPaperID{711} 

\ificcvfinal\fi

\begin{document}

\title{Deep Spatial Pyramid: The Devil is Once Again in the Details}

\author{Bin-Bin Gao \quad Xiu-Shen Wei \quad Jianxin Wu\\
National Key Laboratory for Novel Software Technology\\
Nanjing University, China\\
{\tt\small \{gaobb,weixs\}@lamda.nju.edu.cn, wujx2001@nju.edu.cn}
\and
Weiyao Lin\\
Shanghai Jiao Tong University\\
{\tt\small wylin@sjtu.edu.cn}
}

\maketitle

\begin{abstract}
 In this paper we show that by carefully making good choices for various detailed but important factors in a visual recognition framework using deep learning features, one can achieve a simple, efficient, yet highly accurate image classification system. We first list 5 important factors, based on both existing researches and ideas proposed in this paper. These important detailed factors include: 1) $\ell_2$ matrix normalization is more effective than unnormalized or $\ell_2$ vector normalization, 2) the proposed natural deep spatial pyramid is very effective, and 3) a very small $K$ in Fisher Vectors surprisingly achieves higher accuracy than normally used large $K$ values. Along with other choices (convolutional activations and multiple scales), the proposed DSP framework is not only intuitive and efficient, but also achieves excellent classification accuracy on many benchmark datasets. For example, DSP's accuracy on SUN397 is 59.78\%, significantly higher than previous state-of-the-art (53.86\%).
\end{abstract}

\section{Introduction}

Feature representation is among the most important topics (if not the most important one) in current state-of-the-art visual recognition tasks. Over the past decade, handcrafted features (\emph{e.g.}, SIFT and HOG) were very popular, and they were often encoded into a high dimensional vector by the Bag-of-Visual-Words~(BOVW) framework~\cite{sivic2003video}. The BOVW representation is further improved
by the Vector of the Locally Aggregated Descriptors (VLAD)~\cite{jegou2010aggregating} and Fisher Vector (FV)~\cite{perronnin2010improving} methods, via adding higher order statistics. However, such features are significantly outperformed by the recent deep features from convolutional neural networks (CNNs), which have exhibited significantly better performance than those handcrafted features in visual recognition.

In spite of the impressive results achieved by deep features, there are many factors which can affect the performance of deep feature representations. A lot of factors exist and many details will have huge impact in CNN feature's recognition accuracy. Those factors include, for example, how the deep net is trained. Zhou \emph{et al.}~\cite{zhou2014learning} evaluated deep feature's performance from the same network architecture learned from different training sets (\emph{i.e.}, ImageNet and Places data). They achieved high classification performance on scene recognition tasks with the Places-CNN feature. Chatfield \emph{et al.}~\cite{ChatfieldSVZ14} studied other factors, including architectures of deep nets and data augmentation, \emph{etc}.

After a deep net has been successfully trained, more factors and decisions are awaiting. In other words, \emph{how shall we use the deep features for image recognition?} Studies have been carried out very recently, and some important details have been worked on. However, a systematic study of \emph{``what factors are out there?''} and \emph{``what choices should be made?''} is missing. In this paper, we present our studies to these questions. Specifically, suppose we are given a pre-trained deep CNN model, 
\squishlist
 \item \textbf{What are important factors in utilizing this model?} Based on existing studies in the literature and our new proposals, we make a list of \emph{five important factors}.
 \item \textbf{What decisions are the best concerning these factors?} We carefully evaluate different choices and present our answers to this question. Some choices (\emph{e.g.}, the choice of $K$ size in FV) are quite different from previous practices in the community.
 \item \textbf{What effects do these factors have?} We show that they are \emph{key} to high recognition accuracy. By combining the best choices from the 5 factors we raised, we propose Deep Spatial Pyramid (DSP), a framework that properly utilize deep CNN features. DSP has the following properties:
   \squishlist
    \item[\textbf{*}] \emph{High accuracy.} DSP updates the accuracy of many benchmark datasets in our evaluation. For example, it raises the accuracy of SUN 397 from 53.86\%~\cite{zhou2014learning} to 59.78\%, and Caltech 101 from 93.42\%~\cite{simonyan2014very} to 95.11\%. Note that these previous state-of-the-art results are also based on CNN.
    \item[\textbf{*}] \emph{High efficiency and flexibility.} DSP achieves high processing speed, with roughly 150 ms to process an image. DSP also processes images of any aspect ratio or resolution.
    \item[\textbf{*}] \emph{Small storage cost.} The final DSP representation is memory-efficient, with around 12k dimensions. This length is much shorter than existing combination of CNN features and FV / VLAD, and is advantageous in large-scale problems.
   \squishend
\squishend

We will first present the framework, preliminaries, and the list of important factors in Sec.~\ref{sec:factors}. The study of best decisions for these factors are presented in Sec.~\ref{sec:main}. However, the study of $K$ size is very special, as to have its own Sec.~\ref{sec:K}. DSP is evaluated as a whole system in Sec.~\ref{sec:experiments}, and it is compared with state-of-the-art visual recognition methods. Sec.~\ref{sec:conclusions} concludes this paper.

\section{The framework and important factors} \label{sec:factors}

\begin{figure*}
 \centering
 \includegraphics[width=0.95\textwidth]{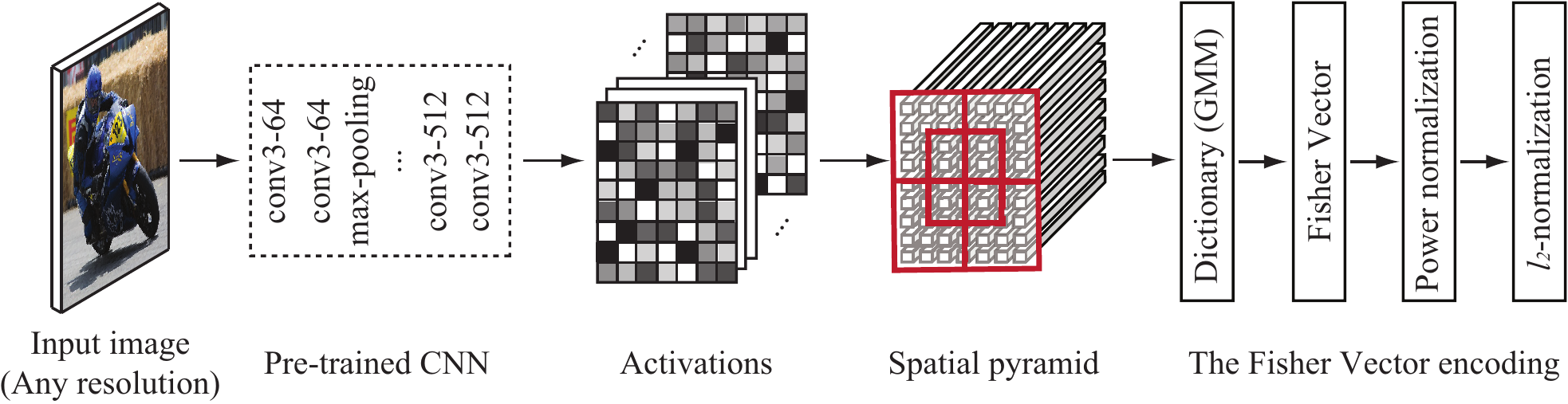}
 \caption{The image classification framework. DSP feeds an arbitrary resolution input image into a pre-trained CNN model to extract deep activations. A GMM visual dictionary is trained based on the deep descriptors from training images. Then, a spatial pyramid partitions the deep activations of an image into $m$ blocks in $N$ pyramid levels. In this way, each block activations are represented as a single vector by the improved Fisher Vector. Finally, we concatenate the $m$ single vectors to form a $2mdK$-dimensional feature vector as the final image representation.} \label{fig:DSP}
\end{figure*}

Our study follow the framework illustrated in Fig.~\ref{fig:DSP}. In the first step, we feed an input image with arbitrary resolution into a pre-trained CNN model to extract deep activations. Then, a visual dictionary with $K$ dictionary items is trained on the deep descriptors from training images. The third step overlay a spatial pyramid partition to the deep activations of an image into $m$ blocks in $N$ pyramid levels.  One spatial block is represented as a vector by using the improved Fisher Vector. Thus, $m$ blocks correspond to $m$ FVs. In the fourth and final step, we concatenate the $m$ FVs to form a $2mdK$-dimensional feature vector as the final image-level representation. 

Our framework does not consider how the pre-trained CNN is obtained or how an image is classified after its representation is obtained. These can be viewed as preliminary factors, and we follow the commonly used decisions for them in the literature.

In practice, some CNN models (\emph{e.g.}, Krizhevsky \emph{et al.}~\cite{krizhevsky2012imagenet} and Zeiler and Fergus~\cite{zeiler2014visualizing}) are popularly used as the deep feature extractor in image related tasks. However, recently neural networks that are even deeper than these are shown to further improve CNN performance, characterized by deeper and wider architectures and smaller convolutional filters when compared to traditional CNN such as ~\cite{krizhevsky2012imagenet,zeiler2014visualizing}. Examples of deeper nets include GoogLeNet~\cite{SzegedyLJSRAEVR14} and VGG Net-D~\cite{simonyan2014very}. Our work is based on the network architecture released by ~\cite{simonyan2014very} (\emph{i.e.}, VGG Net-D). This network consists of 13 layers of $3\times3$ convolutional kernels, with 5 max-pooling layers interspersed, and in the end concluded by 3 fully connected layers. The width of this network starts from 64 in the first layer, increasing by a factor of 2 after each max-pooling layer, until it reaches 512. For the classification, we use a linear SVM classifier.

In the rest of this paper, we follow the notations in~\cite{girshick2014rich}. We use the term ``feature map'' to indicate the convolutional results (after applying the max-pooling) of one filter, the term ``activations'' to indicate feature maps of all filters in a convolutional layer, and the term ``descriptor'' to indicate the $d$-dimensional component vector of activations. ``pool$_5$'' refers to the activations of the max-pooled last convolutional layer, and ``fc$_8$'' refers to the activation of the last fully connected layer.

With these preliminaries and notations, we now discuss the important factors inside this framework.

\squishlist
 \item[1] \emph{Which activation to use?} Deep features for an image can be extracted from either the convolutional layers or the fully connected layers of a pre-trained CNN. The original idea is to use the last fully connected layer directly for classification~\cite{krizhevsky2012imagenet}. And recently, activations from the fully convolutional layers have exemplified its value~\cite{zeiler2014visualizing,LiuSH15,cimpoi2014deep,vi:Xu2015}. Which one shall we adopt?
 \item[2] \emph{How to normalize the deep features} before feeding them into a classifier or the next level of processing? It is not yet a common practice to normalize CNN activations. What are the viable choices and which one is the best?
 \item[3] \emph{How many components in the FV representation?} The GMM model in FV consists of $K$ Gaussian components. It is known that in general a large $K$ (\emph{e.g.}, 256) leads to high accuracy for fully connected activations~\cite{gong2014multi,yoo2014fisher}, dense SIFT~\cite{perronnin2010improving} and action features~\cite{me:Wu2014_CVPRaction}. However, a large $K$ leads to a very long (hundreds of thousands of dimensions) representation. Is a large $K$ really necessary?
 \item[4] \emph{Shall we capture spatial information (and how?)} A general CNN requires a fixed input image size. He \emph{et al.}~\cite{he2014spatial} proposed a SPP-Net to remove the fixed-size constraint, which also inspired a Spatial Pyramid Pooling (SPP). The SPP-Net pooled deep activations of the last convolutional layer and generated fixed length outputs, then the pooled activations were fed into the fully connected layers. Is there a simpler and more natural way to capture spatial information?
 \item[5] \emph{Shall we use information from multiple scales?}  Yoo \emph{et al.}~\cite{yoo2014fisher} replaces the fully connected layers with equivalent convolutional layers to obtain large amount of dense deep descriptors. Then, all the activations are merged into a single vector by Multi-scale Pyramid Pooling (MPP). MPP utilizes multi-scale CNNs' activations. MPP, however, is computationally expensive. Is there an efficient way to capture information from multiple scales?
\squishend

These factors may seem too detailed to be important. However, existing methods adopted very different decisions to these questions, and these differences may well explain their performance differences. We summarize these differences in Table~\ref{table:realtedwork}.

\begin{table}
 \centering
 \caption{Summary of decisions in related methods} \label{table:realtedwork}
 \begin{tabular}{|@{\;}l@{\;}| *{7}{@{\;}c@{\;}|}}
  \hline
  Methods & DF & Resolution & Norm     &  PCA    & K     & SP        & Ms \\ \hline
  SPP-net & C  & fixed      &   -      &  -      & -     & $\surd$   & $\surd$ \\ \hline
  MOP     & F  & fixed      & $\times$ & $\surd$ & 100   & $\times$  & $\surd$ \\ \hline
  MPP     & C  & fixed      & $\times$ & $\surd$ & 256   & $\times$  & $\surd$ \\ \hline
  D-CNN   & C  & any        & $\times$ & $\times$& 64    & $\times$  & $\surd$ \\ \hline \hline
  DSP     & C  & any        & $\surd$  & $\times$& 1,2,3,4  & $\surd$   & $\surd$ \\ \hline
 \end{tabular}
\end{table}

In Table~\ref{table:realtedwork}, ``DF'' refers to deep features, where ``F'' and ``C'' represent the fully connected and convolutional layer, respectively. ``Norm'' refers to how the deep activations are normalized; ``K'' indicates the number of visual words or Gaussian components; ``SP'' refers to spatial pyramid; ``Ms'' refers to multiple scale. In addition, ``-'' means that a method does not involve the corresponding factor. Some methods also use PCA to reduce the dimensionality of deep activations.

From Table~\ref{table:realtedwork}, it is clear that the proposed DSP is flexible (accepting any size image), efficient (fully convolutional and very small $K$), and making full use of the image (spatial pyramid and multiple scales). We will explain how these decisions and choices are made in the next section.

\section{Factors, choices and decisions} \label{sec:main}

We study the 5 factors in this section in Sec.~\ref{sec:fc}--\ref{sec:multi_scale}, respectively. The effect of $K$ size, however, is studied separately in Sec.~\ref{sec:K}.

\subsection{Convolutional vs. fully connected layer} \label{sec:fc}

Convolutional neural networks consist of alternatively stacked convolutional layers and pooling layers, followed by one or more fully connected layers. The convolutional layers generate feature maps by linear convolutional filters with nonlinear activation functions such as rectified linear units, then the feature maps max-pool the outputs within local neighborhoods. Finally, the activations of the last convolutional layer are fed into fully connected layers, followed by a soft-max classifier.

\begin{figure}
 \centering
 \includegraphics[width=0.475\textwidth]{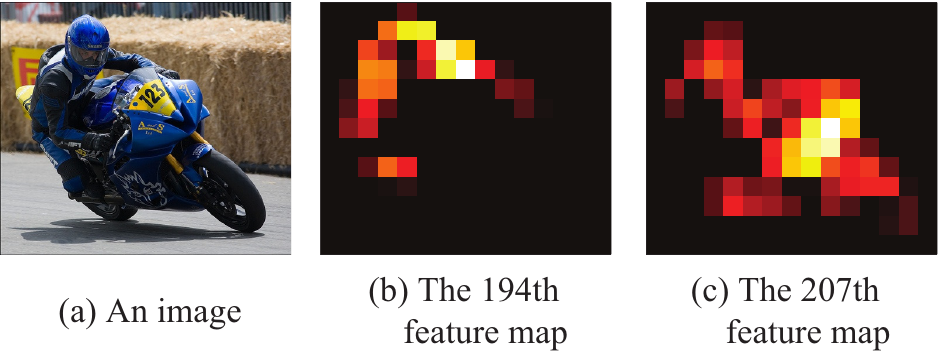}
 \caption{Visualization of the feature maps. (2a) is an image from the PASCAL VOC2007 dataset, (2b) and (2c) are different feature maps of the input image.} \label{fig:fm}
\end{figure}

However, the feature map of top convolutional layers are known to contain mid- and high-level information, \emph{e.g.}, object parts or complete objects~\cite{zeiler2011adaptive}. As shown in Fig.~\ref{fig:fm}, we visualize the input image's feature maps which are generated by the last convolutional layer. In this figure, the strongest response of the 194th and 207th feature map are corresponding to the person and motorcycle in the input image, respectively. Thus, one major difference between convolutional and fully connected layer activations is that the former is directly embedded with rich semantic information of image patches, while the latter not necessarily be so.

Furthermore, the fully connected layers require a fixed image size (\emph{e.g.}, $224\times224$). On the contrary, convolutional layers accept input images of arbitrary resolution or aspect ratio. The pool$_5$ activations can be formulated as a order-3 tensor of size $h\times w \times d$, which include $h \times w$ cells and each cell contains one $d$-dimensional deep descriptor. For example, we will get a $7\times7\times512$ activations if the input image size is $224\times224$. Convolutional layer deep descriptors have been successfully in~\cite{LiuSH15,cimpoi2014deep,vi:Xu2015}.

These deep descriptors contain more spatial information compared to the activation of the fully connected layers, \emph{e.g.}, the top-left cell's $d$-dim deep descriptor is generated using only the top-left part of the input image, ignoring other pixels. In addition, fully connected layers have large computational cost, because it contains roughly 90$\%$ of all the parameters of the whole CNN model. 

Thus, in DSP we use a fully convolutional network by removing the fully connected layers.

\subsection{Normalization and pooling of deep descriptors}

Let $X = [\vec{x}_1, \dots, \vec{x}_t, \dots, \vec{x}_T]^T$ ($X_n\in R^{T\times d}$) be the matrix of $d$-dimensional deep descriptors extracted from an image $I$ via a pre-trained CNN model. $X$ was usually processed by dimensionality reduction methods such as PCA, before they are pooled into a single vector using VLAD or FV~\cite{gong2014multi,yoo2014fisher}. PCA is usually applied to the SIFT features or fully connected layer activations, since it is empirically shown to improve the overall recognition performance. However, our experiments show that PCA significantly hurts recognition when applied to the fully convolutional activations. Thus, it is not applied to fully convolutional deep descriptors in this paper.

In addition, each deep descriptors $\vec{x}_t$ inside $X$ is not normalized in current processing of deep visual descriptors~\cite{cimpoi2014deep}. We first try to normalize $\vec{x}_t$ with the $\ell_2$ vector normalization (\emph{i.e.}, $\vec{x}_t \leftarrow \vec{x}_t / \|\vec{x}_t\|_2$), which leads to better results than null normalization on most datasets, except in \emph{Stanford40}, as shown in Table \ref{table:norm}. 

We also propose a novel $\ell_2$ matrix normalization (\emph{i.e.}, $\vec{x}_t \leftarrow \vec{x}_t / \|X\|_2$), where $\|X\|_2$ is the matrix spectral norm, \emph{i.e.}, largest singular value of $X$. This normalization has a benefit that it normalizes $\vec{x}_t$ using the information from the entire image $X$. It is a bit surprising to observe that it is more effective than the commonly used $\ell_2$ vector normalization, and sometimes by a large margin. An intuitive interpretation is that the $\ell_2$ matrix normalization can use the global information, making it more robust to changes such as illumination and scale.

In order to evaluate the effect of these normalization and PCA for classification performance, we use 4 datasets. We use the original resolution of input images without cropping or warping and pool activations by using FV with $K=4$ (\emph{i.e.}, the GMM has 4 Gaussian components). The experimental results are reported in Table~\ref{table:norm}. The $\ell_2$ matrix normalization before using FV is found to be important for better performance.

\begin{table}
 \centering
 \small
 \caption{Results of the different normalization methods} \label{table:norm}
 \begin{tabular}{|@{\;}l@{\;}| *{4}{@{\;}c@{\;}|}}
  \hline
               & Caltech101 & Stanford40 & Scene15    & Indoor67  \\ \hline
  No           & 90.63      & 74.84      & 90.75      & 71.20 \\
  $\ell_2$ vector & 92.02      & 73.41      & 90.92      & 74.03 \\
  $\ell_2$ matrix & \textbf{92.56}   & \textbf{78.43}    & \textbf{90.99}  & \textbf{74.55} \\
  PCA+$\ell_2$ matrix& 91.95& 75.69      & 90.22      & 71.79  \\ \hline
 \end{tabular}
\end{table}

The size of pool$_5$ is a parameter in CNN because input images have arbitrary sizes. However, the classifiers (\emph{e.g.}, SVM or soft-max) require fixed length vectors. Thus, all the deep descriptors of an image must be pooled to form a single vector. We use the Fisher Vector (FV) to encode the deep descriptors.

We denote the parameters of the GMM with $K$ components by $\lambda = \{\omega_k, \vec{\mu}_k, \vec{\sigma}_k; k=1,\dots,K\}$, where $\omega_k$, $\vec{\mu}_k$ and $\vec{\sigma}_k$ are the mixture weight, mean vector and covariance matrix of the $k^{th}$ Gaussian component, respectively. The covariance matrices are diagonal and $\vec{\sigma}_k^2$ are the variance vectors. Let $\gamma_t(k)$ be the soft-assignment weight of $\vec{x}_t$ with respect to the $k$-th Gaussian, the FV representation corresponding to $\vec{\mu}_k$ and $\vec{\sigma}_k$ are presented as follows~\cite{perronnin2010improving}:
\begin{align}
 \vec{f}_{\vec{\mu}_k}(X) =& \frac{1}{\sqrt{\omega_k}} \sum_{t=1}^T \gamma_t(k) \left( \frac{\vec{x}_t-\vec{\mu}_k}{\vec{\sigma}_k} \right) \,, \\
 \vec{f}_{\vec{\sigma}_k}(X) =& \frac{1}{\sqrt{2\omega_k}} \sum_{t=1}^T \gamma_t(k) \left [\frac{(\vec{x}_t-\vec{\mu}_k)^2}{\vec{\sigma}_k^2}-1 \right ] \,.
\end{align}
Note that, $\vec{f}_{\vec{\mu}_k}(X)$ and $\vec{f}_{\vec{\sigma}_k}(X)$ are both $d$-dimensional vectors. The final Fisher Vector $\vec{f}_\lambda(X)$ is the concatenation of the gradients $\vec{f}_{\vec{\mu}_k}(X)$ and $\vec{f}_{\vec{\sigma}_k}(X)$ for all $K$ Gaussian components. Thus, FV can represent the set of deep descriptors $X$ with a $2dK$-dimensional vector. In addition, the Fisher Vector $\vec{f}_\lambda(X)$ is improved by the power-normalization with the factor of 0.5, followed by the $\ell_2$ vector normalization~\cite{perronnin2010improving}.

We will further study how to choose a proper $K$ size for FV in Sec.~\ref{sec:K}.

\subsection{Deep spatial pyramid}

The proposed method is named as DSP (Deep Spatial Pyramid), since adding spatial pyramid information is the key part of DSP. Adding spatial information through a spatial pyramid~\cite{lazebnik2006beyond} have been shown to significantly improve image recognition performance when dense SIFT features are used. How can we efficiently and effectively utilize the spatial information with fully convolutional activations?

The SPP-net method~\cite{he2014spatial} adds a spatial pyramid pooling layer to deep nets, which has improved recognition performance. However, since we are using FV to pool activations from a fully convolutional network, a more intuitive and natural way exists.

As previously discussed, one single cell (deep descriptor) in the last convolutional layer corresponds to one local image patch in the input image, and the set of all convolutional layer cells form a regular grid of image patches in the input image. This is a direct analogy to the dense SIFT feature extraction framework. Instead of a regular grid of SIFT vectors extracted from $16 \times 16$ local image patches, a grid of deep descriptors are extracted from larger image patches by a CNN.

\begin{figure}
 \centering
 \includegraphics[width=0.6\columnwidth]{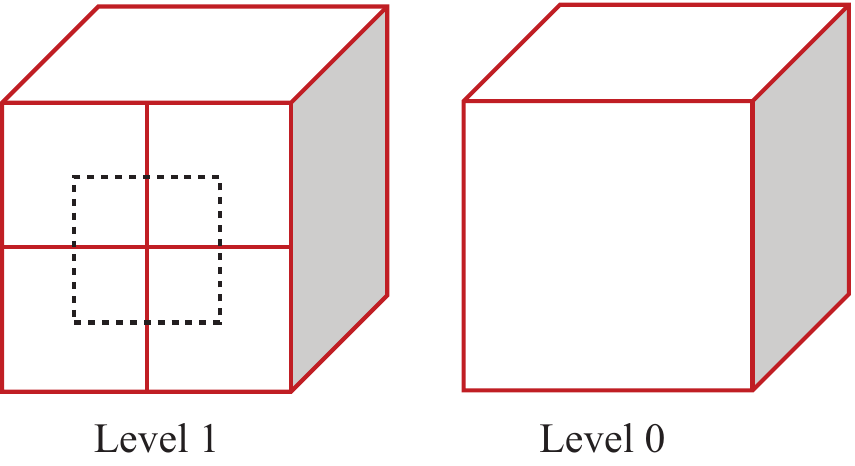}
 \caption{Illustration of the level 1 and 0 deep spatial pyramid.} \label{fig:sp}
\end{figure}

Thus, we can easily form a natural deep spatial pyramid by partitioning an image into sub-regions and computing local features inside each sub-region. In practice, we just need to spatially partition the cells of activations in the last convolutional layer, and then pool deep descriptors in each region separately using FV. The operation of DSP is illustrated in Fig.~\ref{fig:sp}.

The level 0 simply aggregates all cells using FV. The level 1, however, splits the cells into 5 regions according to their spatial locations: the 4 quadrants and 1 centerpiece. Then, 5 FVs are generated from activations inside each spatial region. Note that the level 1 spatial pyramid we use is different from the classic one in~\cite{lazebnik2006beyond}. We follow Wu and Rehg~\cite{wu2011centrist} to use an additional spatial region in the center of the image. A DSP using two levels will then concatenate all 6 FVs from level 0 and level 1 to form the final image representation.

This proposed DSP method is summarized in Algorithm~\ref{algo:DSP}.

\begin{algorithm}
 \caption{The DSP pipeline} \label{algo:DSP}
 \begin{algorithmic}[1]
  \STATE \textbf{Input:}
  \STATE \quad An input image $I$
  \STATE \quad A pre-trained CNN model
  \STATE \textbf{Procedure:}
  \STATE \quad Extract deep descriptors $X$ from $I$ using the\\ \quad pre-defined model, $X = [ \vec{x}_1, \dots, \vec{x}_t, \ldots, \vec{x}_T ]^T$
  \STATE \quad For each activation vector $\vec{x}_t$, perform $\ell_2$ matrix \\ \quad normalization $ \vec{x}_t \leftarrow \vec{x}_t / \|X\|_2 $
  \STATE \quad (Estimate a GMM $\lambda = \{\omega_k, \vec{\mu}_k, \vec{\sigma}_k\}$ using the \\ \quad training set);
  \STATE \quad Generate a spatial pyramid $\{X_1,\dots,X_m\}$ for $X$
  \STATE \quad \textbf{for} all $1\leq i \leq m$
  \STATE \quad ~~~~~~$\vec{f}_{\lambda}(X_i) \leftarrow  [\vec{f}_{\vec{\mu}_1}(X_i),\vec{f}_{\vec{\sigma}_1}(X_i), $ \\
         \quad ~~~~~~~~~~~~~~~~~~~~~~~~~ $\dots,\vec{f}_{\vec{\mu}_K}(X_i),\vec{f}_{\vec{\sigma}_K}(X_i)]$
  \STATE \quad ~~~~~~$\vec{f}_{\lambda}(X_i) \leftarrow \mathop{\mathrm{sign}}(\vec{f}_{\lambda}(X_i)) \sqrt{\vec{f}_{\lambda}(X_i)}$
  \STATE \quad ~~~~~~$\vec{f}_{\lambda}(X_i) \leftarrow \vec{f}_{\lambda}(X_i) / \|\vec{f}_{\lambda}(X_i)\|_2$
  \STATE \quad \textbf{end for}
  \STATE \quad Concatenate $\vec{f}_{\lambda}(X_i)$, $1 \le i \le m$, to form the final \\ \quad spatial pyramid representation $\vec{f}(X)$
  \STATE \quad $\vec{f}(X) \leftarrow \vec{f}(X) / \|\vec{f}(X)\|_2$
  \STATE \textbf{Output:} $\vec{f}(X)$.
 \end{algorithmic}
\end{algorithm}

\subsection{Multi-scale DSP} \label{sec:multi_scale}

In order to capture variations of the activations caused by variations of objects in an image, we generate a multiple scale pyramid, extracted from $S$ different rescaled versions of the original input image. We feed images of all different scales into a pre-trained CNN model and extract deep activations. In each scale, the corresponding rescaled image is encoded into a $2mdK$-dimensional vector by DSP. Therefore, we have $S$ vectors of $2mdK$-dimensions and they are merged into a single vector by average pooling, as
\begin{equation}
 \vec{f}_m = \frac{1}{S}\sum_{s=1}^S \vec{f}_s \,, \label{eq-ap}
\end{equation}
where $\vec{f}_s$ is the DSP representation extracted from the scale level $s$. Finally, $\ell_2$ normalization is applied to $\vec{f}_m$. Note that each vector $\vec{f}_s$ is already $\ell_2$ normalized, as shown in Algorithm~\ref{algo:DSP}.

The multi-scale DSP is related to MPP proposed by Yoo \emph{et al.}~\cite{yoo2014fisher}. A key different between our method and MPP is that $\vec{f}_s$ encodes spatial information while MPP does not.

\section{A small $K$ is better in FV in DSP} \label{sec:K}

\begin{figure*}
 \centering
 \includegraphics[width=0.6\textwidth]{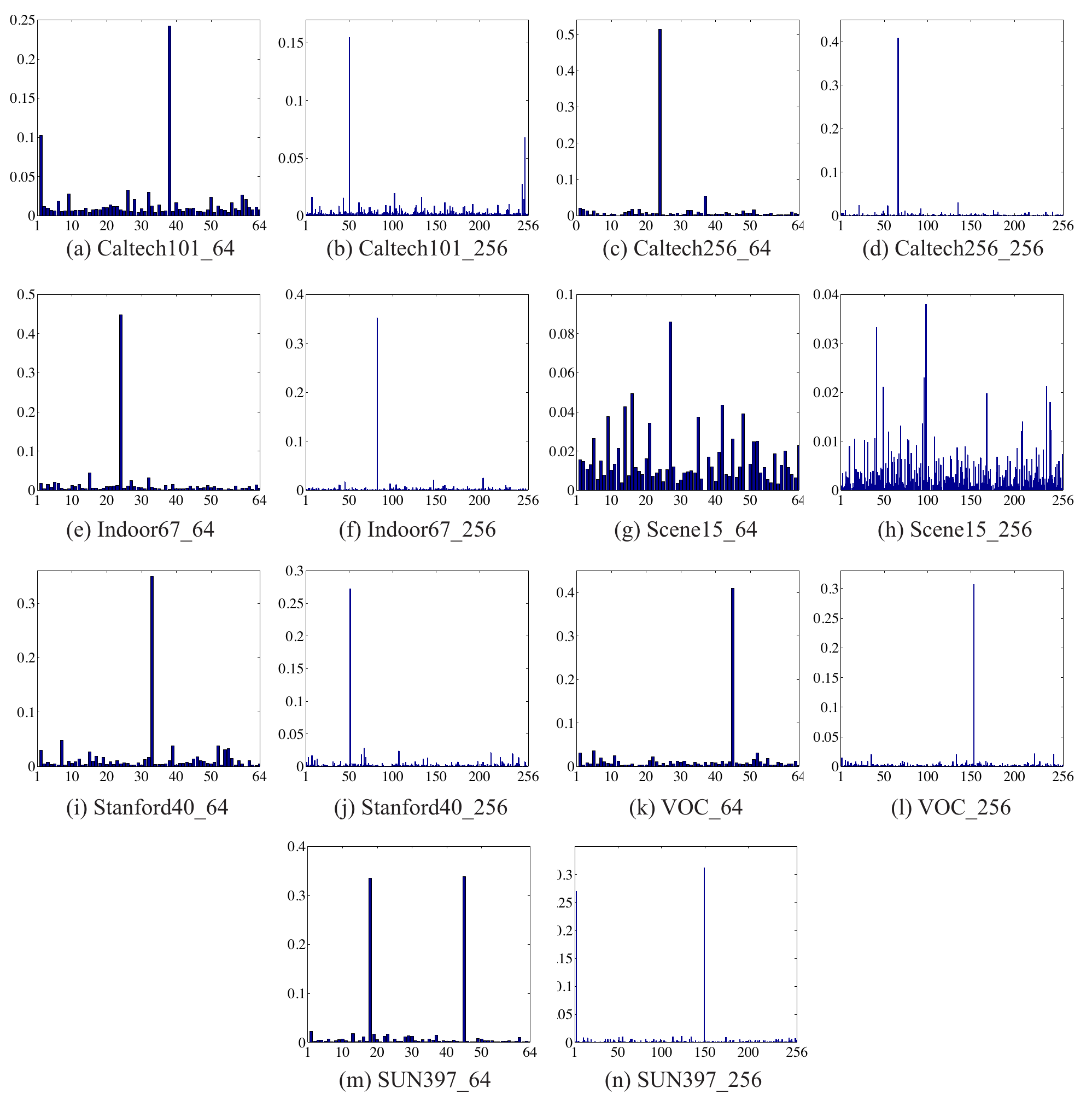}
 \caption{Plot of $\omega$ values in DSP. For each of the seven datasets used in our experiments, we vary the numbers of Gaussian components $K$ to be 64 or 256. (a) and (b) are plots for the \emph{Caltech-101} data set, with $K$ being 64 and 256, respectively. The meaning of other plots can be deduced from their captions similarly. Note that, the plots for \emph{Scene15} are not similar to other plots. When $K$ is larger than 4, DSP could achieve satisfactory classification accuracy rates in Scene 15, a trend that is consistent with the plots shown in (g) and (h).} \label{fig:w}
\label{fig:short}
\end{figure*}

\begin{figure*}
 \centering
 \subfloat[\emph{Caltech-101} and \emph{Scene15}]
 {
   \includegraphics[width=0.67\columnwidth]{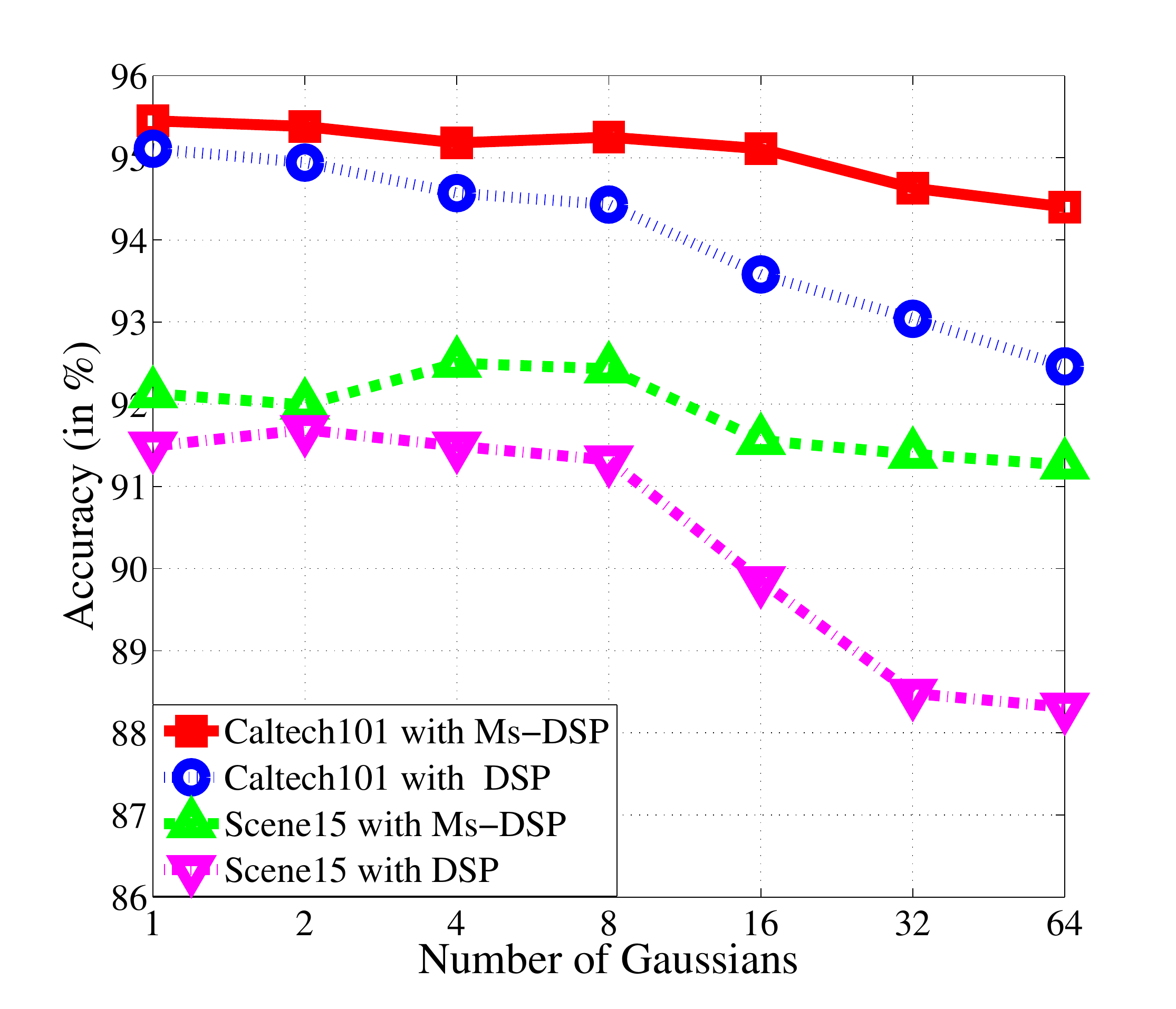} \label{fig:a}
 }
 \subfloat[\emph{Stanford40} and \emph{Indoor67}]
 {
   \includegraphics[width=0.67\columnwidth]{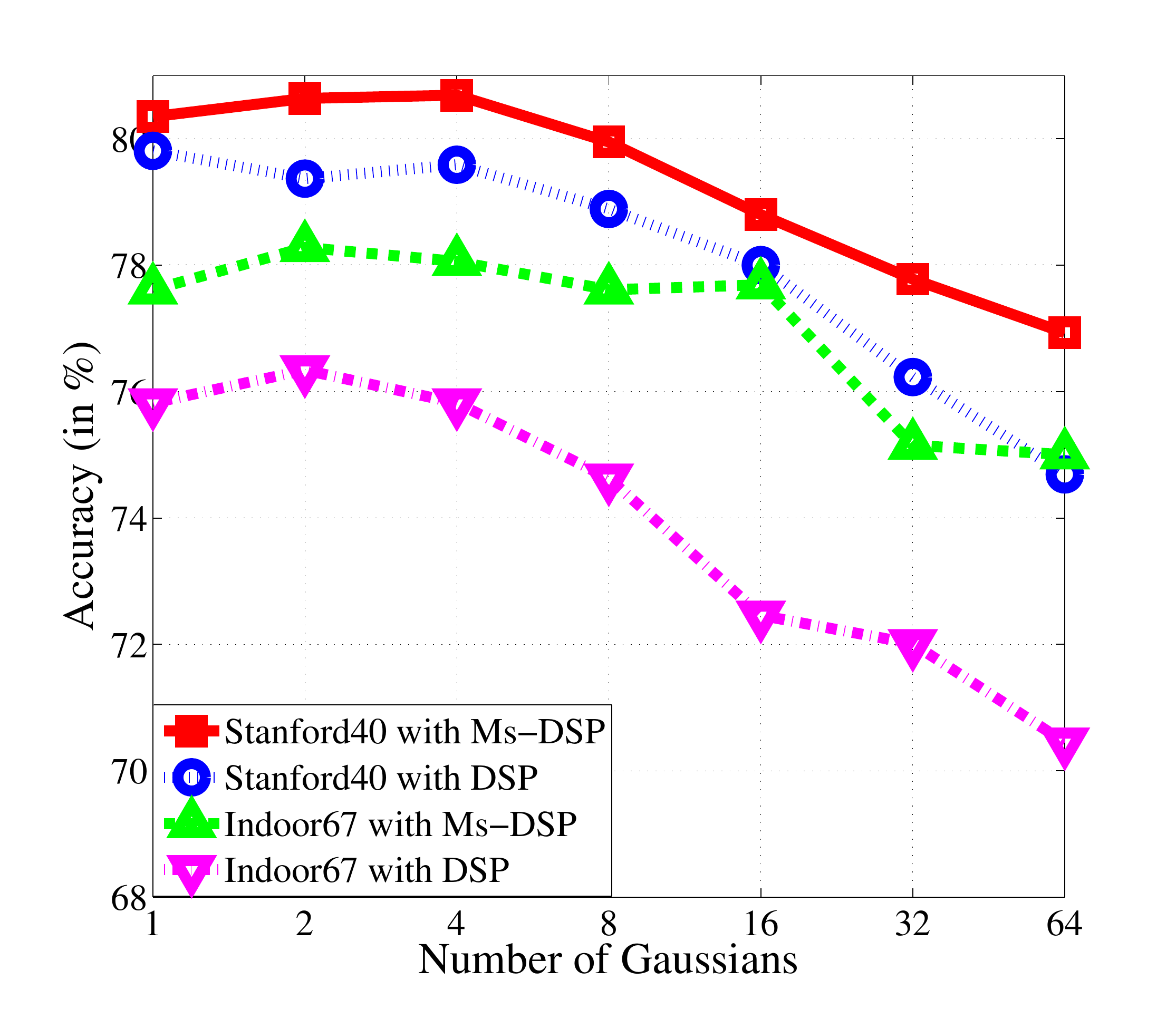} \label{fig:b}
 }
 \caption{Classification performance of DSP and Ms-DSP with different numbers of Gaussians} \label{fig:GMM}
\end{figure*}

In this section, we will discuss one key character of DSP, \emph{i.e.}, the number of GMM's components.

Our experiments show that in DSP, when the number of GMM's components $K$ is small (\emph{e.g.}, from 1 to 4), it will achieve satisfactory classification performances. In fact, when different $K$ are used, the highest recognition accuracy is usually achieved by setting $K$ to 1 or 2!

This phenomenon is not consistent with common practices in image classification by using local descriptors via the FV encoding. When deep learning features are used together with FV, a large $K$ value is also used. Moreover, Yoo \emph{et al.}~\cite{yoo2014fisher} specified the value of $K$ to be 256 when they trained their visual vocabulary. More previous examples of large $K$ values can be found in Table~\ref{table:realtedwork}. Having a small $K$ value is very beneficial in terms of CPU and storage costs, however, why is DSP requiring a small $K$?

We believe the answer is because DSP uses a small number of deep descriptors per image, \emph{i.e.}, $h \times w$ is a small integer. We usually extract no more than 100 512-dimensional deep descriptors from the last convolutional layer from one image, while~\cite{yoo2014fisher} represented one image with 4,410 vectors of 4,096 dimensional dense CNN activations. If the value of $K$ is specified as a large number (\emph{e.g.}, 128 or 256), the resulting FV representation will be problematic.

First, if a large $K$ is used in DSP, there will not be enough deep descriptors to estimate an accurate GMM model, because each training image will only contribute few number of deep descriptors. An inaccurate GMM model will adversely affect the classification performance seriously. Second, many FV components will only contain zeros, because there are more Gaussian components than CNN descriptors. We conjecture that this will cause FV to lose accuracy.

We also empirically study this phenomenon. As shown in Fig.~\ref{fig:w}, we plot distribution of GMM components' priors (\emph{i.e.}, $\omega$) in DSP. There are 14 plots for the 7 datasets used in our experiments. Two plots are shown for each data set, which corresponds to different number of GMM components (shown as the horizontal axis), \emph{i.e.}, 64 and 256. The vertical axis shows the value of $\omega$ for each Gaussian component.

It is obvious to find that: for most datasets, one or two $\omega$ values are much larger than the rest. For example, when $K=64$ in the SUN 397 dataset, the two tall bars indicate that two $\omega$ values are above 0.3, and their sum is around 0.7. In other words, only 2 Gaussian components are responsible for more than 70\% of the variations of the distribution. The rest 30\% might be related to noisy or background image patches. Thus, $K=2$ might be the best choice in this particular case. In most datasets, we can observe the same phenomenon: one or two Gaussian components are dominating the entire distribution. This observation might explain why DSP just needs a small number of Gaussian components. Since a small value of $K$ in DSP will cause a much lower computational cost, it is efficient to handle large scale image classification tasks.

We further evaluate the impact of $K$ in DSP and multiple scale DSP (Ms-DSP). We show the classification results in Fig.~\ref{fig:GMM} as a function of the number of Gaussians (\emph{i.e.}, $K$) of the GMM, and $K$ is increased by a factor of 2. A smaller $K$ (\emph{e.g.}, $K=2$) always obtains better classification performance for DSP and Ms-DSP. With the increasing of $K$, we can see that DSP and Ms-DSP lead to a drop in the discriminative ability. DSP or Ms-DSP feature vector may be too sparse when $K$ is increased, which is detrimental to classification. When $K=2$, a DSP representation has only $2 \times 512 \times 2 \times 6 = 12288$ dimensions. The entire DSP pipeline (from reading in an image till emitting a prediction) requires on average 0.15 second per image.

For a fixed $K$, Ms-DSP always significantly outperforms DSP. This is not surprising since, for a given $K$, Ms-DSP captures more information from rescaled images, which DSP does not have access to.

\section{Experiments} \label{sec:experiments}

The purpose of this section is to evaluate the performance of DSP as a complete pipeline. We report results in three object recognition datasets, \emph{Caltech-101}~\cite{fei2007learning}, \emph{Caltech-256}~\cite{griffin2007caltech} and \emph{Pascal VOC 2007}~\cite{pascal2007}, and three scene recognition datasets, \emph{Scene15 categories}~\cite{lazebnik2006beyond}, \emph{MIT Indoor67}~\cite{quattoni2009recognizing} and \emph{SUN397}~\cite{sanchez2013image}, and one action recognition data set, \emph{Stanford40}~\cite{yao2011human}. Except for \emph{Pascal VOC 2007} and \emph{MIT Indoor67} which have fixed training and test splittings, all experiments on the other datasets are repeated as the average of three randomly sampled train/test splittings.

\subsection{Datasets}

\emph{Caltech-101}~\cite{fei2007learning} contains 9K labeled images of 101 object categories and a background category. We follow the procedure of ~\cite{fei2007learning} and randomly select 30 images per category for training and test on up to 50 images per class in every split. \emph{Caltech-256}~\cite{griffin2007caltech} with 31K images and 257 classes is an improvement of \emph{Caltech-101}. Following ~\cite{griffin2007caltech}, each split contains 60 training images per class and the rest is used for test. For \emph{PASCAL VOC 2007} which contains 20 object classes, we use its standard protocol and measure the average precision (AP) and report the mean AP (mAP) of 20 categories.

\emph{Scene15} is composed of 15 different kinds of scenes, where each category has 200 to 400 images. We randomly select 100 images per class for training and the rest for test, following~\cite{zhou2014learning}. \emph{MIT Indoor67}~\cite{quattoni2009recognizing} is a challenging indoor data set comparing with outdoor scene recognition. The dataset has 15,620 images with 67 indoor scene categories. The standard split~\cite{quattoni2009recognizing} for this dataset consists of 80 training and 20 test images per category. \emph{SUN397}~\cite{xiao2010sun} is the largest data set for scene recognition. It contains 397 categories and each category has at least 100 images. The training and test splits are fixed and publicly available from~\cite{xiao2010sun}, where each split has 50 training and 50 test images per category. We select the first three splits from the 10 public splits in our experiments.

\emph{Stanford40}~\cite{yao2011human} contains 40 diverse daily human actions and with 180$\sim$300 images for each category. In each splitting, we randomly select 100 images in each class for training and the remaining for test.

In our experiments, average accuracy rate is used to evaluate the classification performances on \emph{Caltech-101}, \emph{Caltech-256}, \emph{MIT Indoor67}, \emph{Scene15}, \emph{SUN397}, and \emph{Stanford40}. For \emph{PASCAL VOC 2007}, we employ mean average precision (mAP) to evaluate our proposed method and other approaches.

\subsection{Experiment details}

In our DSP,  VGG Net-D~~\cite{simonyan2014very} is employed as the pre-trained CNN model to extract deep activations.  For simplicity, pre-trained CNN model weights are kept fixed without fine-tuning. Note that, we just employ VGG Net-D without its fully connected layers in our experiments, thus can accept input images of arbitrary sizes. Input images do not need to be resized into a fixed aspect ratio. However, considering running efficiency, an image is resized such that the smallest and largest edge of input image will not be lower than 224 or higher than 1120, respectively. In addition, each image is preprocessed by subtracting the per-pixel mean (of the ImageNet images and provided along with the CNN model). 

We use $K=2$ in FV in this section. An image is represented by the concatenation of FVs from all the $6$ sub-blocks in a two level deep spatial pyramid. For using multi-scale, the rescaled images are $s$ times of the of original input image, where $s \in \{1.4,1.2,1.0,0.8,0.6\}$. the FVs of all five scale are merged into a single vector by average pooling as Eq.~\ref{eq-ap}.

One-versus-rest linear SVM is used for classification. Following~\cite{zhou2014learning}, all classifiers use the same parameters $C=1$ for fair comparisons. Our experiments use the following open source libraries: VLFeat~\cite{vedaldi08vlfeat}, MatConvNet~\cite{VedaldiMatConvNet} and LIBLINEAR~\cite{fan2008liblinear}.

\subsection{Main results}

\begin{table*}[t]
 \centering
 \caption{Recognition accuracy (or mAP) comparisons on seven datasets. The highest accuracy (mAP) of each column is marked in bold. \cite{simonyan2014very}'s results were achieved using VGG Net-D and VGG Net-E, evaluation was measured by mean class recall on \emph{Caltech-101}, \emph{Caltech-256} instead of accuracy .} \label{table:main_results}
 \small
 \begin{tabular}{|l|l|ccccccc|}
  \hline
  Methods &Description & Caltech-101 & Caltech-256  &VOC 2007 &Scene15 &SUN397 &MIT Indoor67 &Stanford40 \\ \hline \hline
  \multirow{6}{*}{SoA} & \cite{he2014spatial} & \textbf{93.42$\pm$0.50}  & -  & 82.44 & - & -  & -  & - \\
  & \cite{gong2014multi} & - & - & - & - & 51.98 & 68.88 & - \\
  & \cite{yoo2014fisher} & - & - & 82.13 & - & - & 77.56 &-  \\
  & \cite{zhou2014learning}  & 84.79$\pm$0.66 & 65.06$\pm$0.25 & - & 91.59$\pm$0.48  & 53.86$\pm$0.21 & 70.80 & 55.28$\pm$0.64 \\
  & \cite{ChatfieldSVZ14} & 88.35$\pm$0.56 & 77.61$\pm$0.12 & 82.4 & -  & - & - & - \\
  & \cite{simonyan2014very} & 92.7$\pm$0.5 (*) & \textbf{86.2$\pm$0.3}(*) & \textbf{89.7} &-  &- &- &-\\ \hline \hline
  Baseline
  & Fc$_8$       & 90.55$\pm$0.31 & 82.02$\pm$0.12 & 84.61 & 89.88$\pm$0.76 & 53.90$\pm$0.45 & 69.78 & 71.53$\pm$0.34 \\ \hline \hline
  &Pool$_5$+FV   & 90.03$\pm$0.75 & 79.48$\pm$0.53 & 88.12 & 89.00$\pm$0.42 & 51.39$\pm$0.51 & 71.57 & 73.96$\pm$0.52 \\ 
  \multirow{3}{*}{Our}
  & DSP          & 94.66$\pm$0.26 & 84.22$\pm$0.11 &88.60  & 91.13$\pm$0.77 & 57.27$\pm$0.34 & 76.34 & 79.75$\pm$0.34 \\
  & Ms-DSP       &\textbf{95.11$\pm$0.26} & \textbf{85.47$\pm$0.14} & \textbf{89.31} & \textbf{91.78$\pm$0.22} & \textbf{59.78$\pm$0.47} & \textbf{78.28} & \textbf{80.81$\pm$0.29} \\ \hline
 \end{tabular}
\end{table*}

\begin{table*}[t]
 \centering
 \caption{Per-class classification performance on PASCAL VOC 2007.} \label{table:voc}
 \footnotesize
 \begin{tabular}{|@{\,}l@{\,}|@{\,}l@{\,}| *{20}{@{\,}c@{\,}}|}
  \hline
  Methods &Description  	&	 aero 	&	 bike 	&	 bird 	&	 boat 	&	 bottle 	&	 bus 	 &	 car 	 &	 cat 	 &	 chair 	&	 cow 	&	 table 	&	 dog 	&	 horse 	 &	 mbike 	&	 person 	&	 plant 	 &	 sheep 	 &	 sofa 	 &	 train 	&	 tv \\
  \hline\hline
  Baseline
  &Fc$_8$ &	96.27 	&	90.81 	&	93.81 	&	92.40 	&	58.24 	&	86.01 	&	90.92 	 &	91.91 	&	69.45 	 &	 78.08 	&	79.36 	&	90.87 	&	91.69 	&	88.98 	&	95.35 	&	61.31 	&	 88.14 	&	71.68 	&	 96.53 	&	 80.28\\ \hline \hline
  &Pool$_5$+FV & 97.23& 94.44& 96.12& 93.54& \textbf{70.99}& 88.45& 93.43& 95.48& 71.16& 81.33& 82.21& 93.55& 95.08& 90.51& \textbf{97.64}& 69.84& 88.70& 77.42& 96.92& \textbf{88.29}\\
  \multirow{3}{*}{Our}
  &DSP &	97.45 	&	94.12 	&	96.79 	&	\textbf{94.98} 	&	69.64 	&	87.99 	&	93.28 	&	95.76 	&	 72.75 	&	81.65 	&	85.07 	&	94.31 	&	94.84 	&	91.57 	&	97.53 	&	69.61 	&	89.42 	&	80.14 	 &	97.47 	&	87.64\\
  &Ms-DSP &	\textbf{97.67} 	&	\textbf{95.24} 	&	\textbf{96.84} 	&	94.47 	&	70.58 	&	\textbf{89.32} 	&	 \textbf{93.50} 	&	\textbf{95.92} 	&	\textbf{74.61} 	&	\textbf{83.99} 	&	\textbf{85.68} 	&	\textbf{95.27} 	 &	\textbf{95.37} 	&	\textbf{92.02} 	&	97.42 	&	\textbf{71.05} 	&	\textbf{90.82} 	&	\textbf{80.5}7 	&	 \textbf{97.69} 	&	88.14\\
  \hline
 \end{tabular}
\end{table*}

State-of-the-art and two baseline results are reported in Table~\ref{table:main_results}. In particular, the first baseline method is fc$_8$ which is extracted from the last fully connected layer. To extract fc$_8$ feature, we resize the image so that its resolution is $224\times224$. $\ell_2$-normalization is applied to the fc$_8$ activations before employing SVM, which was suggested in~\cite{ChatfieldSVZ14}. The other baseline is the pool$_5+$FV where deep descriptors are aggregated to single vector by orderless FV pooling. In order to compare fairly, we use the same resolution of input image as in our DSP.

On most datasets, fc$_8$ already performs well. Pool$_5$ produces quite good results even though the Pool$_5$ activations are computed using only 10$\%$ of the CNN parameters of the complete CNN model, which shows that fully convolutional features (with small $K$ in FV and $\ell_2$ matrix normalization) are powerful, especially on \emph{VOC2007} (84.61$\%\rightarrow$ 88.12$\%$) and \emph{Stanford40} (71.53$\%\rightarrow$ 73.96$\%$).

DSP and multi-scale DSP can significantly outperform baseline and state-of-the-arts methods. Compared to the baselines, DSP improves performance in all datasets by 1--5$\%$, especially on \emph{SUN397} (53.90\%$\rightarrow$59.27\%) and \emph{Stanford40} (73.96\%$\rightarrow$79.75\%). This gain is mainly due to the fact that DSP can capture the spatial information on top of pool$_5$ activations. On the other hand, the fully convolutional network relaxes the constraint that the input images must have the same fixed size, thus the full image can be fed into a pre-trained CNN without changing its aspect ratio. Combining multiple scale and DSP (Ms-DSP) achieves the best recognition performance on all datasets. Since fully convolutional and small $K$ are used, Ms-DSP is still very efficient.

Our DSP and Ms-DSP can achieve mean recall $96.38\pm0.53$ and $96.88\pm0.59$ on \emph{Caltech-101}, respectively, and $90.05\pm0.07$ and $90.89\pm0.17$ on \emph{Caltech-256}, respectively. These results are significantly higher than that of~\cite{simonyan2014very} (92.7\% for \emph{Caltech-101} and 86.2\% for \emph{Caltech-256}).

In addition, on the VOC2007 dataset, our best performance is slightly lower ($0.4\%$) than that in~\cite{simonyan2014very}. However, \cite{simonyan2014very} used fusion feature which was computed using two pre-trained CNN (\ie, VGG Net-D and VGG Net-E). Detailed VOC results in Table~\ref{table:voc} show that our methods are better than fc$_8$ in every category.

\section{Conclusion} \label{sec:conclusions}

In order to present a powerful deep feature representation, details have to be made right. In other words, decisions for important factors must be carefully studied and made. In this paper, we picked a list of 5 important factors and provided our answers to them. The main findings of this paper form a complete pipeline DSP (deep spatial pyramid), which integrates the following components: activations from the last convolutional layer, naturally processing input image of any size instead of fixed size, dense deep features extracted from multiple scales, and most importantly, a natural way to build a spatial pyramid in deep learning. DSP, in spite of being simple and efficient, has excellent performance in many benchmark datasets.

In particular, we emphasize the following new findings.
\squishlist
 \item Normalization: $\ell_2$ matrix normalization is more effective than unnormalized or $\ell_2$ vector normalization.
 \item DSP: DSP can effectively capture the spatial information in a natural and efficient manner.
 \item $K$ size in FV: Pooling deep descriptors only need small number of Gaussian components in the Fisher Vector, which leads to lower computational costs.
\squishend

Other factors and details can be further considered in the DSP framework, which we will study in the future. For example, convolutional activations from multiple layers (cross-layer~\cite{LiuSH15}) might further improve classification accuracy. And VLAD might be a better fit than FV for aggregating deep convolutional activations~\cite{vi:Xu2015}.

{\small
\bibliographystyle{ieee}
\bibliography{egbib}
}

\end{document}